\documentclass[sigconf]{acmart}

\usepackage{amsmath, amsfonts, amsthm, xspace, color}
\usepackage{booktabs} 

\usepackage{multirow, tabularx} 
\usepackage{enumitem}
\usepackage{flushend}
\usepackage[T1]{fontenc}
\usepackage{epstopdf}
\usepackage{balance}
\usepackage{graphicx}
\usepackage{url}
\usepackage{wrapfig,lipsum}
\usepackage{makecell}
\usepackage{wrapfig}
\usepackage{latexsym}

\usepackage{microtype}
\newcommand\BibTeX{B\textsc{ib}\TeX}
\usepackage{caption}
\usepackage{subcaption}

\newcommand{\hide}[1]{} 

\newcommand{\etc}{\emph{etc.}\xspace} 
\newcommand{\ie}{i.e.\xspace} 
\newcommand{\eg}{e.g.\xspace} 
\newcommand{\nop}[1]{}
\newcommand{\mquote}[1]{{``\emph{#1}''}}

\newtheoremstyle{exampstyle}
  {0pt} 
  {0pt} 
  {\itshape} 
  {1em} 
  {\bfseries} 
  {.} 
  {.5em} 
  {} 

\theoremstyle{exampstyle}
\newtheorem{thm:def}{Definition}

\newtheorem{thm:eg}{Example}
\newtheorem{thm:lem}{Lemma}
\newtheorem{thm:obs}{Observation}
\newtheorem{thm:req}{Requirement}
\newtheorem{thm:prop}{Proposition}
\newtheorem{thm:principle}{Principle}
\newtheorem{thm:thm}{Theorem}
\newtheorem{thm:corollary}{Corollary}

\newcommand{\pair}[1]{$\langle #1 \rangle$}			

\newcommand{\Our}{\mbox{\textsf{ExHalder}}\xspace}
\newcommand{\OurNoPT}{\mbox{\textsf{ExHalder-NoPT}}\xspace}
\newcommand{\OurNoEX}{\mbox{\textsf{ExHalder-NoEX}}\xspace}
\newcommand{\OurNoHC}{\mbox{\textsf{ExHalder-NoHC}}\xspace}

\usepackage{bbm}
\usepackage{adjustbox}
\usepackage[super]{nth}

\AtBeginDocument{%
  \providecommand\BibTeX{{%
    \normalfont B\kern-0.5em{\scshape i\kern-0.25em b}\kern-0.8em\TeX}}}

\setcopyright{acmcopyright}
\copyrightyear{2023}
\acmYear{2023}
\setcopyright{rightsretained}
\acmConference[WWW'23]{Proceedings of the ACM Web Conference 2023}{May 1--5, 2023}{Austin, TX, USA}
\acmBooktitle{Proceedings of the ACM Web Conference 2023 (WWW'23), May 1--5, 2023, Austin, TX, USA}
\acmDOI{10.1145/3543507.3583375}
\acmISBN{978-1-4503-9416-1/23/04}

\begin{document}

\title{``Why is this misleading?'': Detecting News Headline \\ Hallucinations with Explanations}

\author{Jiaming Shen}
\affiliation{%
  \institution{Google Research}
  \country{}
}
\email{jmshen@google.com}

\author{Jialu Liu}
\affiliation{%
  \institution{Google Research}
    \country{}
}
\email{jialu@google.com}

\author{Dan Finnie}
\affiliation{%
  \institution{Google}
    \country{}
}
\email{danfinnie@google.com}

\author{Negar Rahmati}
\affiliation{%
  \institution{Google}
    \country{}
}
\email{negarr@google.com}

\author{Michael Bendersky}
\affiliation{%
  \institution{Google Research}
    \country{}
}
\email{bemike@google.com}

\author{Marc Najork}
\affiliation{%
  \institution{Google Research}
  \country{}
}
\email{najork@google.com}

\renewcommand{\shortauthors}{Jiaming Shen et al.}

\begin{abstract}

Automatic headline generation enables users to comprehend ongoing news events promptly and has recently become an important task in web mining and natural language processing.
With the growing need for news headline generation, we argue that the \emph{hallucination issue}, namely the generated headlines being not supported by the original news stories, is a critical challenge for the deployment of this feature in web-scale systems
Meanwhile, due to the infrequency of hallucination cases and the requirement of careful reading for raters to reach the correct consensus, it is difficult to acquire a large dataset for training a model to detect such hallucinations through human curation.
In this work, we present a new framework named \Our to address this challenge for headline hallucination detection.
\Our adapts the knowledge from public natural language inference datasets into the news domain and learns to generate natural language sentences to explain the hallucination detection results.
To evaluate the model performance, we carefully collect a dataset with more than six thousand labeled $\langle$article, headline$\rangle$ pairs.
Extensive experiments on this dataset and another six public ones demonstrate that \Our can identify hallucinated headlines accurately and justifies its predictions with human-readable natural language explanations.
\end{abstract}



\maketitle
\section{Introduction}\label{sec:intro}

With tens of millions of news articles published every day on the web~\cite{news_volume}, people are inundated with massive news contents and find them hard to digest.
To facilitate more efficient and user-friendly news content consumption, recent works in the industry propose to generate headlines from either a single news article~\cite{Ao2021PENSAD} or a set of news articles related to the same event~\cite{Gu2020GeneratingRH}.
The generated news headline is intended to serve as a succinct, informative, and accurate summary of its underlying news article(s), and thus it helps the users to quickly grasp the gist of a news story.

\begin{figure}[!t]
    \centering
    \centerline{\includegraphics[width=0.48\textwidth]{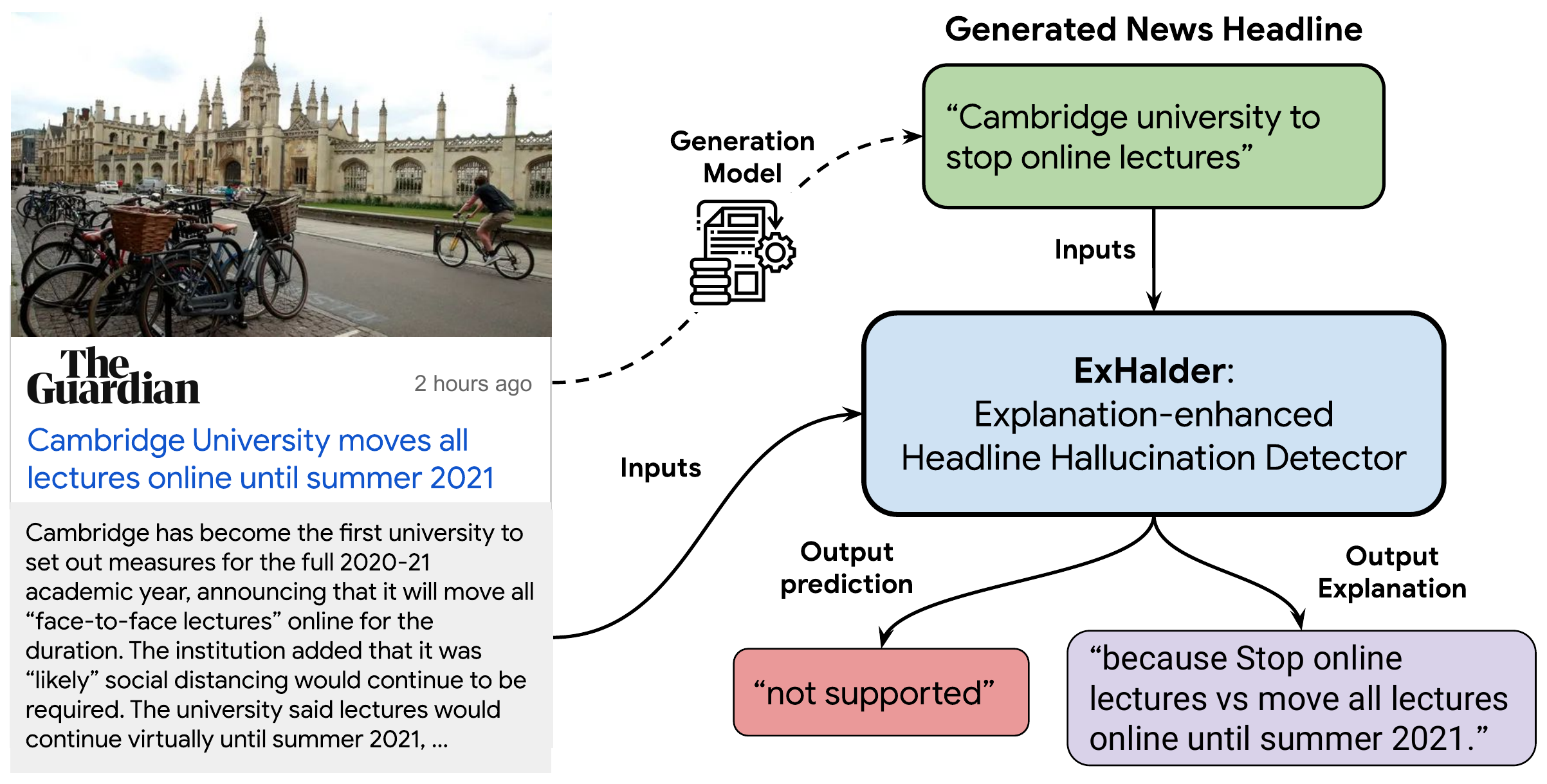}}
    \caption{An illustrative example of automated news headline hallucination detection with a model generated natural language explanation.}
    \label{fig:intro_example}
\end{figure}

To obtain high-quality news headlines, early studies~\cite{Mohammadzadeh2012TitleFinderET, Kochetkova2018NewsHA, Higurashi2018ExtractiveHG} propose \emph{extractive} methods to first extract words from the article title and then organize those salient words into the output headline.
More recently, with the advances of natural language generation research~\cite{Rush2015ANA, Tang2022RecentAI, Zhang2020PEGASUSPW}, more \emph{abstractive} methods are developed to directly summarize the news article into a concise news headline~\cite{Tan2017FromNS, Bukhtiyarov2020AdvancesOT, Gu2020GeneratingRH, Ao2021PENSAD, Li2022NewsHG}.
These abstractive summarization methods typically adopt the encoder-decoder architecture~\cite{Cho2014LearningPR, Sutskever2014SequenceTS} where the encoder synthesizes the knowledge in the news article using vector representations and the decoder outputs the generated headline in a word-by-word fashion.
Although overall quality improvements have been made by this approach, people observe that these generation models often will output hallucinated headlines that are not supported by the underlying news articles.
For example, in Figure~\ref{fig:intro_example}, the generation model outputs the headline \mquote{Cambridge university to stop online lectures} based on the article with the title ``Cambridge University moves all lectures online until summer 2021''.
The generated headline is misleading because it suggests that Cambridge University will stop online lectures instead of moving some face-to-face lectures online until the summer of 2021.

In this paper, we study the \emph{news headline hallucination detection} task: given a pair of $\langle$news article, news headline$\rangle$, we aim to algorithmically determine if the headline is supported by the underlying article and thus is not misleading.
Figure~\ref{fig:intro_example} shows an example where the news article indicates Cambridge University will move in-person lectures online for a period of time but the generated news headline suggests the opposite.
Therefore, this is a misleading headline and the hallucination detector should predict this headline as ``not supported''.
An intuitive approach to this task is to train a classifier using a large set of $\langle$article, headline$\rangle$ pairs with their hallucination labels.
However, as those hallucination cases appear infrequently and require deep reading comprehension, such a labeled dataset is usually of small scale and thus forbids us from learning a powerful model that can capture the subtle semantic differences between news articles and news headlines.

To tackle the lack-of-supervision challenge, we propose a novel framework named \textbf{\Our}, standing for ``\textbf{\underline{Ex}}planation-enhanced Headline \textbf{\underline{Hal}}lucination \textbf{\underline{de}}tecto\textbf{\underline{r}}''.
\Our is developed based on two key ideas.
First, we observe that there exist many similarities between the headline hallucination detection (HHD) task and the natural language inference (NLI)~\cite{maccartney2009natural, Bowman2015ALA} task.
For example, both of them aim to detect if one piece of text (``headline'' in the HHD task and ``hypothesis'' in the NLI task) is supported/entailed by another piece of text (``article'' in the HHD task and ``premise'' in the NLI task).
Based on this observation, we propose to pretrain \Our using public large-scale NLI datasets~\cite{Bowman2015ALA, Camburu2018eSNLINL, Yin2021DocNLIAL} and transfer the knowledge learned from the NLI task to the headline hallucination detection task.
Second, as the framework name suggests, we propose to go beyond the binary class label and utilize natural language explanations to augment the model learning process.
These explanations are particularly useful in the low resource setting (\ie, with limited training data) and help models to generalize better.
We demonstrate that the learned \Our can generate high-quality human-readable explanations to justify its prediction results.
Take the case in Figure~\ref{fig:intro_example} for example, \Our not only predicts the headline is ``not supported'' by the news article but also justifies the output with an explanation ``because Stop online lectures vs move all lectures online until summer 2021''.

To make the best use of these explanations, \Our includes three key components: 
(1) a \emph{reasoning classifier} which receives as input the $\langle$article, headline$\rangle$ pair and outputs the class label along with the label explanation,
(2) a \emph{hinted classifier} which receives as input the $\langle$article, headline, explanation$\rangle$ triplet and predicts the class label, and
(3) an \emph{explainer} that generates the natural language explanation based on the input $\langle$article, headline$\rangle$ with its known class label.
These three components utilize the explanation signals from different angles and work collaboratively within our \Our framework.
Specifically, during the training phase, we will train the explainer to generate more explanations and use them to augment the original training set for learning the reasoning classifier and the hinted classifier. 
At the inference stage, we first input the test $\langle$article, headline$\rangle$ tuple into the reasoning classifier to obtain its predicted class and generated explanation.
Then, we concatenate the explanation with the input tuple and feed them together into the hinted classifier to obtain another class prediction.
Finally, we aggregate these two predictions and return the final predicted class with its corresponding explanation.

We test the effectiveness of the \Our framework on seven hallucination detection datasets from different domains.
Our results demonstrate that \Our achieves state-of-the-art performance in terms of detection accuracy, recall, and F1 score.
Furthermore, we show that \Our can generate high-quality natural language explanations to justify its prediction results.

\begin{figure*}[!t]
    \centering
    \centerline{\includegraphics[width=0.98\textwidth]{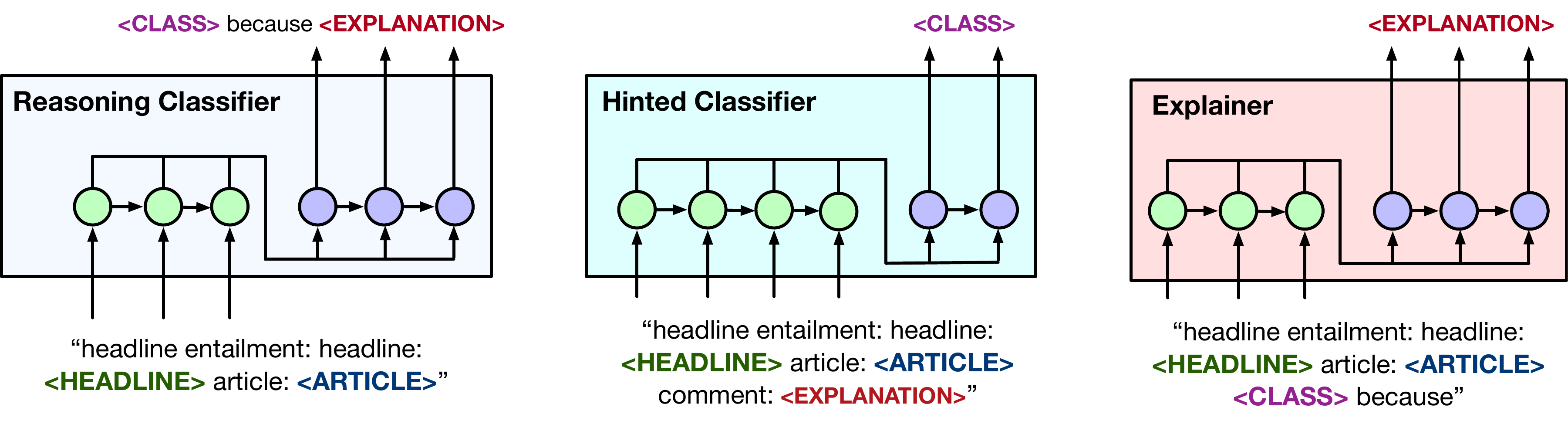}}
    \caption{Key Components of \Our framework.}
    \label{fig:framework_component}
\end{figure*}

\smallskip
\noindent \textbf{Contributions}.
To summarize, our major contributions include: 
(1) a novel framework that automatically detects news headline hallucinations with limited manually labeled data;
(2) an effective method for integrating natural language explanations into the detection pipeline and enabling the model to generate human-readable explanation;
(3) a real-world headline hallucination detection datasets curated by news-domain experts; and
(4) extensive experiments on seven real-world datasets that verify both the hallucination detection accuracy and the generated explanation quality.

\smallskip
The rest of the paper is organized as follows. 
Section~\ref{sec:related_work} discusses the related work. 
Section~\ref{sec:problem} formalizes our problem.
Then, we present our \Our framework in Section~\ref{sec:method} and conduct experiments in Section~\ref{sec:experiment}. 
Finally, we conclude this paper in Section~\ref{sec:conclusion}.


\section{Related Work}\label{sec:related_work}

\noindent \textbf{News Headline Generation.}~
Automated news headline generation, widely considered as a special form of document summarization task, aims to generate a headline-style summary from either a single news article~\cite{Murao2019ACS, Ao2021PENSAD} or a set of news articles related to the same event~\cite{Gu2020GeneratingRH}.
Early studies address this task by adopting an extractive approach~\cite{Mohammadzadeh2012TitleFinderET, Kochetkova2018NewsHA} that first selects words from the article and then organizes them into the output headline via statistical models~\cite{Banko2000HeadlineGB, Schwartz2002AutomaticHG, Dorr2003HedgeTA}. 
This approach achieves limited success as some extracted words are incoherent~\cite{Alfonseca2013HEADYNH} and the traditional statistical models lack expressive powers to generate vivid text.
Recently, the advances of natural language generation research~\cite{Rush2015ANA, Tang2022RecentAI, Zhang2020PEGASUSPW} lead to more abstractive headline generation methods~\cite{Gavrilov2019SelfAttentiveMF, Gu2020GeneratingRH, Bukhtiyarov2020AdvancesOT, Ao2021PENSAD, Li2022NewsHG}.
They adopt the encoder-decoder architecture~\cite{Cho2014LearningPR, Sutskever2014SequenceTS} where the encoder synthesizes the knowledge in the news article(s) using vector representations and the decoder outputs the generated headline in a word-by-word fashion with potential constraints (\eg, length control~\cite{Kikuchi2016ControllingOL}, keyword preservation~\cite{Mao2020ConstrainedAS}, or style preference~\cite{Xu2019ClickbaitSH}).
Although the overall quality improvements have been made, people observe that these generation models often will output hallucinated headlines that are not supported by the underlying news articles~\cite{Xiao2021OnHA, Ji2022SurveyOH}.
This hallucination issue becomes a key blocker for deploying web-scale automated headline generation models in industry, which motivates us to study the news headline hallucination detection problem in this work.

\smallskip
\noindent \textbf{Hallucination Detection.}~
Recent years have witnessed the great improvements of many natural language generation (NLG) models.
One remaining challenge for deploying these NLG models in real-world systems is the hallucination issue that refers to the scenario where the generated content being nonsensical or unfaithful to the provided source content~\cite{Magdy2010WebbasedSF,Filippova2020ControlledHL,Ji2022SurveyOH}.
Many studies propose to mitigate the hallucination issue by either cleaning the model training data~\cite{Gardent2017CreatingTC,Rebuffel2021ControllingHA} or learning a classifier to postprocess/filter generated contents~\cite{Cao2020FactualEC,Chen2021ImprovingFI,Rashkin2021IncreasingFI}.
In a boarder sense, our study falls into the second category and further enhances the classifier with a natural language explanation component.

\smallskip
\noindent \textbf{Natural Language Inference.}
The task of natural language inference~\cite{Bowman2015ALA} (also called textual entailment~\cite{Dagan2005ThePR, BarHaim2006TheSP}) aims to predict if a given ``premise'' text entails, contradicts, or is neutral with regard to another ``hypothesis'' text.
As this task can measure the model's language reasoning capability and has multiple large datasets~\cite{Bowman2015ALA, Williams2018ABC, Yin2021DocNLIAL}, there have been studies on how to adapt it for other language tasks such as weakly-supervised classification~\cite{Shen2021TaxoClassHM}, sentence embedding learning~\cite{Conneau2017SupervisedLO}, and fact checking~\cite{Sathe2021AutomaticFW}.
Among these studies, the most relevant are those utilizing trained NLI models for measuring the faithfulness of summarization methods~\cite{FEVER, MNBM, FRANK}.
However, different from this work, they do not leverage the explanation information. In contrast, our experiments show that these explanations can help better transfer the knowledge from the NLI task to the headline hallucination detection task.

\smallskip
\noindent \textbf{Natural Language Explanation.}
Leveraging natural language explanations to improve machine learning models has long been studied in the literature. 
Typical usages include feeding the human-written explanations as additional input signals~\cite{Liang2020ALICEAL, Hase2020LeakageAdjustedSC} or treating them as model outputs and training the model to reproduce them~\cite{Narang2020WT5TT}.
Although how models benefit from these explanations still remains an active research problem~\cite{Hase2022WhenCM}, the general finding is that these natural language explanations could be particularly useful when only limited amount of labeled data are provided~\cite{Lee2020LEANLIFEAL, Ye2020TeachingMC, Wei2022ChainOT}.
In this work, we study how to effectively leverage these explanations to enhance the hallucination detection accuracy and explore the possibility of generating free-text explanations to justify model's reasoning rationale.

\section{Problem Formulation}\label{sec:problem}

In this section, we first introduce the notations used later in the paper and then present our problem formulation.

\smallskip
\noindent \textbf{Notations.}
A news \emph{article} $\mathbf{d} \in \mathcal{D}$ is a document composed of a token sequence $[d_1, d_2, \dots]$.
A news \emph{headline} $\mathbf{h} \in \mathcal{H}$ is a succinct summary of the news article, represented by another token sequence $[h_1, h_2, \dots]$.
Although the news headline is typically generated based on the news article, those generation methods sometimes encounter the \emph{hallucination} issue, namely the generated headline is not entailed by its corresponding news article.
Given a pair of article and headline \pair{\mathbf{d}_i, \mathbf{h}_i}, we use $s_i \in \mathcal{S}=\{0, 1\}$ to indicate if the headline is supported by the article or not.
Optionally, we may have a natural language explanation to elucidate why the article supports or contradicts the headline.
We use another token sequence $\mathbf{e} = [e_1, e_2, \dots]$ to denote this free-text explanation.


\smallskip
\noindent \textbf{Problem Definition.}
The task of \emph{news headline hallucination detection} is to learn a predictor $\mathbf{f}(\cdot): \mathcal{D} \times \mathcal{H} \rightarrow \mathcal{Y}$ that takes a pair of $\langle$news article, news headline$\rangle$ as the input and predicts if the news headline is supported by the news article.
Based on the available resources for learning the predictor $\mathbf{f}(\cdot)$, we further consider two settings: (1) \textbf{supervised setting} where we have a small set of $N$ labeled examples $\{\mathbf{d}_i, \mathbf{h}_i, s_i\}|_{i=1}^{N}$ to train the predictor, and (2) \textbf{zero-shot setting} where we do not have any labeled example and have to exploit knowledge from other related tasks.


\section{\Our: Explanation-Enhanced Headline Hallucination Detector}\label{sec:method}
In this section, we first introduce three key components of our \Our framework.
Then, we elaborate on how \Our utilizes these components for news headline hallucination detection and how to train the \Our framework.
Finally, we discuss the inference procedure of \Our framework.

\begin{figure*}[!t]
    \centering
    \centerline{\includegraphics[width=0.98\textwidth]{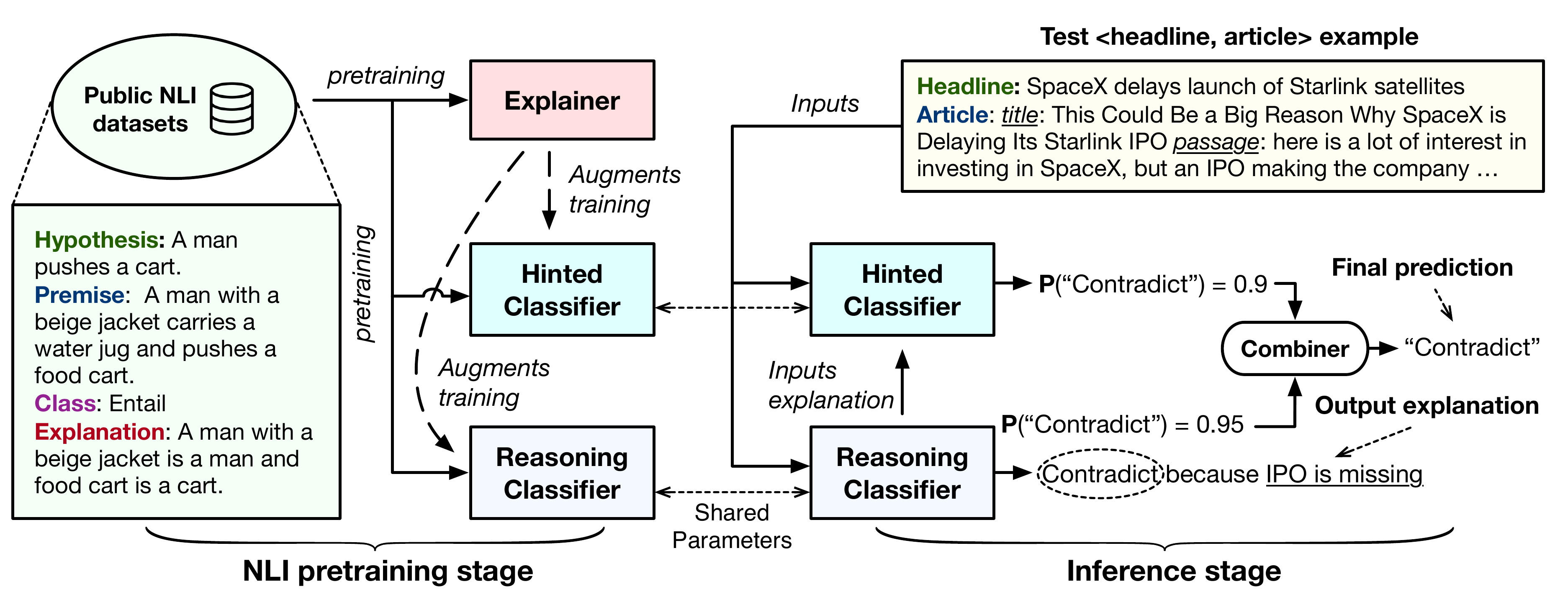}}
    \caption{The \Our framework overview.}
    \label{fig:framework}
\end{figure*}

\subsection{Key Components of \Our Framework}

In this work, we adopt the widely used encoder-decoder architecture~\cite{Cho2014LearningPR, Sutskever2014SequenceTS} due to its strong representation power and wide applicability for both classification and generation tasks.
The encoder first compresses the information of an input sequence $\mathbf{x} = [x_1, x_2, \dots]$ into its vector representation and then the decoder generates tokens in the output sequence $\mathbf{y} = [y_1, y_2, \dots]$ one at a time.
Specifically, given the full input sequence $\mathbf{x}$ and the output sequence prefix $\mathbf{y}_{1:i-1} = [y_1, y_2, \dots, y_{i-1}]$, we produce the token $y_i$ as follow:
\begin{equation}
    \mathbf{P}(y_i|\mathbf{x}, \mathbf{y}_{1:i-1}) = \frac{\exp( \mathbf{v}_i \cdot \mathbf{E}(y_i))}{\sum_{y' \in \mathcal{V}} \exp(\mathbf{v}_i) \cdot \mathbf{E}(y')) },
\end{equation}
where $\mathbf{v}_i$ is the decoder output hidden vector corresponding to token $y_i$, $\mathbf{E}(y)$ is the embedding of a token $y$, and $\mathcal{V}$ denotes the entire vocabulary.

Although initially proposed for generation tasks, the encoder-decoder models can also be applied to classification problems by (1) choosing one special token for each possible class; (2) forcing the model to do one-step decoding; and (3) mapping the output token $y_1$ to its corresponding class as the final prediction~\cite{t5}.
Take our hallucination detection task as an example.
We can use the special token `$c$' in vocabulary $\mathcal{V}$ to represent the ``contradictary'' class and compute the hallucination probability as $\mathbf{P}(y_1 = \text{`}c\text{'}|\mathbf{x})$.

In the \Our framework, given a pair of $\langle$article, headline$\rangle$ ($\langle \mathbf{d}_i$, $\mathbf{h}_i \rangle$) along with its class label $s_i$ and label explanation $\mathbf{e}_i$, we define the following three components based on how we construct their input sequences $\mathbf{x}$ and expect the output sequences $\mathbf{y}$ will be. 
Figure~\ref{fig:framework_component} shows an architecture overview of these three components.

\smallskip
\noindent \textbf{Reasoning Classifier.}
The input sequence $\mathbf{x}$ is of format ``headline entailment: headline: <HEADLINE> article: <ARTICLE>'' where the ``<HEADLINE>'' and ``<ARTICLE>'' are two placeholders and will later be replaced with the contents in the news headline $\mathbf{h}$ and the news article $\mathbf{d}$.
The output sequence $\mathbf{y}$ is of format ``<CLASS> because <EXPLANATION>'' where the placeholder token ``<CLASS>'' (one of \{``Entail, ``Contradict''\}) indicates if the news article entails or contradicts the news headline, and the placeholder token ``<EXPLANATION>'' corresponds to the natural language explanation $\mathbf{e}$.
When the rater does not provide any explanation for the labeled example during the curation process, this ``<EXPLANATION>'' token could simply be an empty string.
Note here if we throw away the ``because <EXPLANATION>'' part in the output sequence $\mathbf{y}$, the reasoning classifier will degenerate into a standard classifier with the encoder-decoder architecture.

\smallskip
\noindent \textbf{Hinted Classifier.}
As its name suggests, the input of hinted classifier goes beyond the one used for the reasoning classifier and includes the natural language explanation as the ``hint''.
Specifically, we append a string ``comment: <EXPLANATION>'' after the reasoning classifier's input and teach the model to output a single token ``<CLASS>'' to indicate the final predicted class.
The hinted classifier is expected to achieve better classification performance than the reasoning classifier because (1) its input contains more signals from the additional ``comment: <EXPLANATION>'' part, and (2) it does not waste representative power for the explanation generation.

\smallskip
\noindent \textbf{Explainer.}
Different from the previous two ``classifiers'', the explainer inputs a sequence that already contains the class information and aims to output a natural language sentence to explain this class.
Specifically, the input sequence of the explainer is of format ``headline entailment: headline: <HEADLINE> article: <ARTICLE> <CLASS> because'' and the output sequence will be just the natural language explanation itself.

\subsection{The \Our Framework}
Our \Our framework is built upon the above three key components for news headline hallucination detection.
As both the reasoning classifier and the hinted classifier contain the prediction result ``<CLASS>'' in their outputs, one may argue that we can directly adopt supervised learning techniques to train these two classifiers for hallucination detection.
This approach, however, requires massive labeled data which are often inaccessible for real-world applications.
Therefore, in this work, we propose two novel techniques to address such a label data scarcity issue: (1) pretraining with large-scale natural language inference (NLI) datasets, and (2) augmented training with human-written explanations.
Figure~\ref{fig:framework} shows an overview of our \Our framework.
 
\subsubsection{NLI-based Pretraining.}
The natural language inference (NLI) task aims to predict if a given ``hypothesis'' is supported/entailed by another input ``premise'' text.
Take the case in Figure~\ref{fig:framework} as an example, the hypothesis \mquote{A man pushes a cart} is supported by the premise \mquote{A man with a beige jacket carries a water jug and pushes a food cart.} and thus the target class is ``Entail''.
We observe that this NLI task shares many similarities with our news headline hallucination detection (HHD) task.
Both of them aim to detect if one piece of text (``headline'' in the HHD task and ``hypothesis'' in the NLI task) is supported/entailed/grounded by another piece of text (``article'' in the HHD task and ``premise'' in the NLI task).
Such a connection enables us to transfer knowledge from the NLI task to our news domain HHD task.
Furthermore, different from the case in the news domain HHD with limited labeled data, there are many large-scale publicly available NLI datasets~\cite{Bowman2015ALA,Williams2018ABC, Camburu2018eSNLINL, Nie2020AdversarialNA, Yin2021DocNLIAL}.

Based on the above observation, in this work, we propose to pretrain all the components in \Our using the NLI datasets.
Specifically, we use the eSNLI~\cite{Camburu2018eSNLINL} and ANLI~\cite{Nie2020AdversarialNA} datasets for pretraining as they both contain human written natural language explanations. 
Given a NLI example $\langle$hypothesis, premise, label$\rangle$, we first construct one training example by replacing the ``<HEADLINE>'' and the ``<ARTICLE>'' placeholder tokens with the ``hypothesis'' text and the ``premise'' text, respectively.
Then, we train our reasoning classifier, hinted classifier, and explainer models using the standard teacher-forcing technique~\cite{Williams1989ALA}.

\subsubsection{Explainer-augmented Training.}
Due to language variability, people have different ways to express the same underlying rationale.
However, in the existing NLI datasets, due to constrained manual curation resources, each example has only a very limited amount of human-written explanation(s) (\eg, 1 for the eSNLI dataset and 1-3 for the ANLI dataset).
To obtain more explanations and use them to train the hinted classifier and the reasoning classifier, we propose to augment the existing NLI datasets with a learned explainer.
Specifically, after the initial pretraining stage, we use the learned explainer to generate $K$ additional explanations for each NLI example.
Then, we merge these augmented examples with the examples in the original NLI dataset and continue to train the hinted classifier and reasoning classifier with this augmented dataset.
More training details are discussed in the experiment section.

\subsubsection{Optional Domain Fine-tuning.}
For both the NLI-based pretraining step and the explainer-augmented training step, we only use the general domain datasets.
When additional news domain-specific datasets are available, we can follow the same procedure above and further fine-tune the components in our \Our framework.
In this work, we collect a new headline hallucination dataset and perform this domain fine-tuning step in one of our experiment settings (c.f. Section~\ref{subsec:supervised_setting}).

\subsection{\Our Inference}
At the inference stage, we are given a test $\langle$article, headline$\rangle$ pair and apply the learned hinted classifier and reasoning classifier to make a prediction.
Specifically, we first feed the test example into the reasoning classifier and parse its output sequence into the predicted class and the explanation sentence.
Then, we concatenate this generated explanation with the original headline and article and treat it as the input sequence of the hinted classifier.
We use the hinted classifier to obtain another class prediction.
Finally, we use a combiner to aggregate the predictions from the reasoning classifier and the hinted classifier.
Here, without requiring more labeled examples, we adopt a simple averaging strategy for the combiner. 
Namely, we average the probability scores from the reasoning classifier and the hinted classifier and return this averaged score as the final prediction probability\footnote{\small When more labeled examples are available, another combiner design is to train a small model to calibrate and aggregate the probability scores from both the reasoning classifier and the hinted classifier.}.


\section{Experiments}\label{sec:experiment}
In this section, we study the performance of \Our on two settings: (1) \emph{supervised setting} where we have a small set of labeled $\langle$article, headline$\rangle$ pairs for model learning, and (2) \emph{zero-shot setting} where no labeled data is provided. 

\subsection{News Headline Hallucination Detection with Supervision}\label{subsec:supervised_setting}

\subsubsection{Dataset.}
To the best of our knowledge, there is no publicly available news headline hallucination detection dataset.
Therefore, in this paper, we collect a new dataset that contains 6270 human curated examples: 5190 examples for training, 349 examples for validation, and 731 examples for testing.
Each example includes a triplet of $\langle$news article, news headline, hallucination label$\rangle$ where the headline is generated from NHNet~\cite{Gu2020GeneratingRH} and the label is obtained from multiple human experts according to a common guideline.
Specifically, we ask three full-time journalism degree holders in the news domain to rate each example and determine the final hallucination label through majority voting.
Among these examples, 1934 of them are labeled as "hallucinated" and the remaining 4336 examples are labeled as "entailed". 
Furthermore, there are 2074 examples with additional rater-written comments (besides binary hallucination labels) and we treat them as user-provided explanations.
The dataset is publicly available at: \url{https://bit.ly/exhalder-dataset}.

\subsubsection{Compared Methods.}
We compare the following methods for the headline hallucination detection task:
\begin{itemize}[leftmargin=*]
    \item SVM~\cite{Cortes2004SupportVectorN}: We manually extract a set of features based on the textual string of the news headline and the news article (\eg, their corresponding sequence lengths, the number of overlapping words, some word-level  editing distances like Jaro-Winkler distance~\cite{Cohen2003ACO}, \etc), and train a standard SVM model with the RBF kernel for predictions.
    \item XGBoost~\cite{Chen2016XGBoostAS}: Similar to the above SVM method, we feed those handcrafted features to the standard XGBoost classification model for detecting the hallucinations.
    \item BERT$_{base}$~\cite{Devlin2019BERTPO}: We concatenate the headline and the article text (with a [SEP] separator) and feed it into the pretrained BERT base model for prediction.
    \item T5$_{xxl}$~\cite{t5}: Similar to BERT, we input the concatenated headline and article to the encoder module of T5 and use its decoder to output one single token indicating the final predicted class.
    \item T5$_{xxl}$ + Exp: We incorporate the natural language explanation information into the T5$_{xxl}$ model by requiring its decoder to output the class token followed by the explanation. This is similar to the reasoning classifier architecture in our \Our framework.
    \item \OurNoPT: Our \Our framework without the NLI-based pretraining step. Namely, we just train the explainer, the reasoning classifier, and the hinted classifier on our news-domain hallucination detection training set.
    \item \OurNoEX: Our \Our framework with the NLI-based pretraining step but without leveraging any explanation information. Namely, we force the reasoning classifier to just output one token indicating the hallucination label and remove the hinted classifier as well as the explainer components.
    \item \OurNoHC: Our \Our framework without the hinted classifier module. Namely, we only train the explainer and the reasoning classifier during the pretraining stage and use the reasoning classifier alone for prediction at the inference stage.
    \item \Our: The full version of our proposed framework.
\end{itemize}
We implement SVM using scikit-learn, XGBoost using its official codebase\footnote{\small \url{https://xgboost.readthedocs.io/en/stable/index.html}.}, and BERT$_{base}$ method using the Tensorflow Model Garden\footnote{\small \url{https://github.com/tensorflow/models/tree/master/official/nlp}.}.
For T5$_{xxl}$, T5$_{xxl}$+Exp, and \Our along with its variants, we develop them based on the T5X library\footnote{\small \url{https://github.com/google-research/t5x}.} and use the T5-11B model in the following experiments. 
More implementation details and hyper-parameter settings are discussed in Appendix~\ref{app:exp_details}.

\subsubsection{Experiment Settings.}
As we formulate the headline hallucination detection as a classification problem, we adopt the standard classification evaluation metrics: Accuracy, Precision, Recall, and F1 score. 
Among these metric, we emphasize that the recall value indicates the percentage of hallucinated headlines captured by the hallucination detector.
Better recall means less misleading headlines will be surfaced to users and thus leads to more positive user experiences.
For each tested method, we run it for five times and report the averaged results.
Finally, for performance comparisons, we we conduct statistical significance test using the two-tailed paired $t$-test with 95\% confidence level.

\begin{table}[!t]
\centering
\scalebox{1.0}{
\begin{tabular}{l|cccc}
\toprule
Methods & Accuracy & Precision & Recall & F1  \\
\midrule
SVM             & 57.31 & 28.65 & 20.50 & 23.90 \\
XGBoost         & 60.19 & 42.39 & 60.67 & 49.91 \\
\midrule
BERT$_{base}$     & 73.46 & 71.43 & 31.38  & 43.60  \\
T5$_{xxl}$        & 82.39 & 76.29 & 66.93 & 71.29  \\
T5$_{xxl}$ + Exp    & 82.62 & 78.98 & 64.15 & 70.63  \\
\midrule
\OurNoPT    & 82.08  & 75.96 & 66.11 & 70.69 \\
\OurNoEX    & 83.17  & 80.01$^{*}$ & 64.71 & 71.54  \\
\OurNoHC    & 84.08$^{*}$  & 82.06$^{*}$ & 65.69 & 72.96$^{*}$  \\
\Our        & \bf 84.46$^{*}$ & \bf 82.63$^{*}$ & \bf 67.16$^{*}$ & \bf 74.08$^{*}$  \\
\bottomrule
\end{tabular}
}
\vspace*{0.2cm}
\caption{Quantitative results on the news headline hallucination detection dataset. The superscript $^{*}$ means the improvement is statistically significant compared to T5$_{xxl}$.}
\label{table:hallucination_main_results}
\vspace*{-1.5em}
\end{table}

\subsubsection{Experiment Results.}
Below we first present the main experiment results and compare \Our with the baseline methods.
Then, we conduct ablation analysis to study how the key components of \Our impact the framework overall performance.
Finally, we present a few case studies to demonstrate the potential impacts of \Our in real-world scenarios.

\smallskip
\noindent \textbf{1. Overall Detection Performance.}
Table~\ref{table:hallucination_main_results} presents the results of all compared methods.
First, we can see that the results of those traditional methods with manual feature engineering (\ie, SVM, XGBoost) are unsatisfactory.
This shows that headline hallucination detection is a challenging task and requires models to capture the subtle semantic differences between the article and the headline.
Second, we compare \Our with \OurNoPT and see that the NLI-based pretraining indeed helps us to better identify the hallucinated headlines by warming up the model with entailment task semantics.
Third, by comparing \Our with \OurNoEX, we observe further performance improvements and this demonstrates that injecting the explanation information into the model training process is useful.
Finally, we can see our proposed \Our has the overall best performance across all the metrics and defeats the second-best method by a large margin.

\begin{figure}[!t]
    \centering
    \centerline{\includegraphics[width=0.48\textwidth]{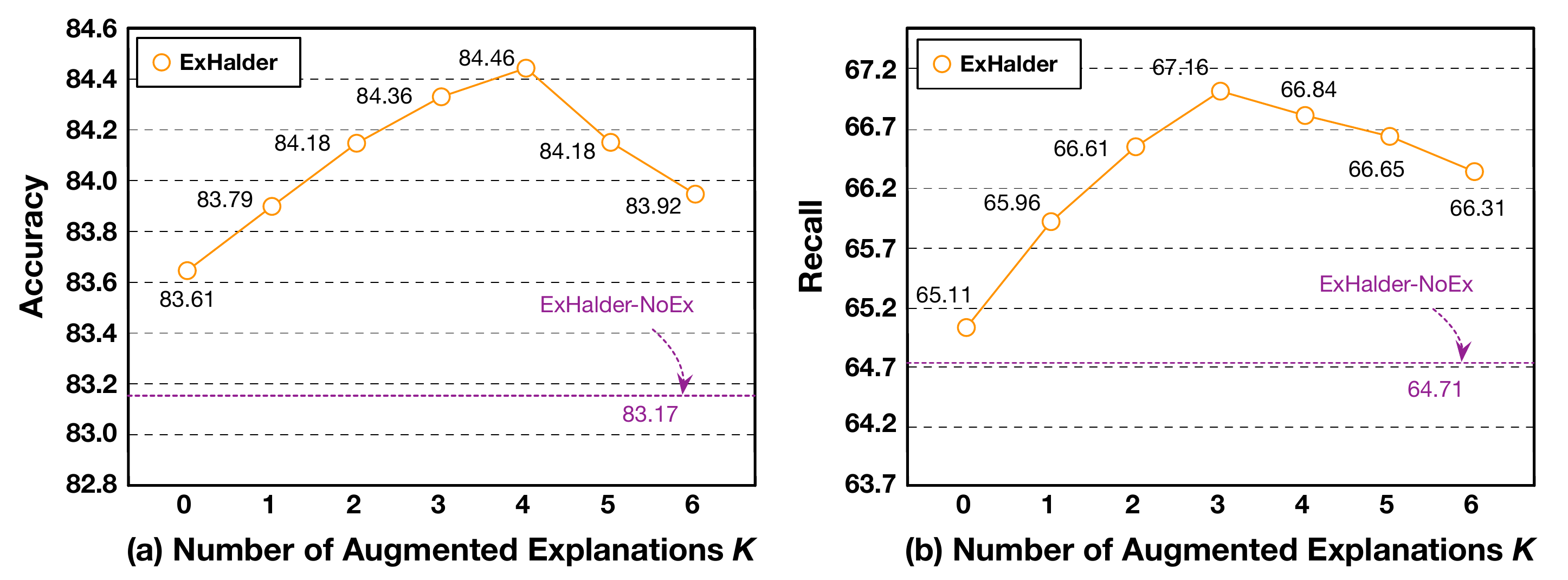}}
    \caption{Parameter sensitivity analysis on the news hallucination detection dataset. We vary the number of explanations generated by the explainer component and compute the accuracy and recall of \Our.}
    \label{fig:ablation_study_k}
    \vspace*{-1.5em}
\end{figure}

\smallskip
\noindent \textbf{2. Ablation Analysis of Model Components.}
\Our contains three key components: a reasoning classifier, a hinted classifier, and an explainer.
The above \OurNoEX demonstrates the importance of the reasoning classifier component and the explainer component.
Here, we study how the hinted classifier components affect the performance of \Our.
As shown in Table~\ref{table:hallucination_main_results}, we can see that removing the hinted classifier leads to low prediction accuracy and significantly hurts the hallucination detection recall. 

\smallskip
\noindent \textbf{3. Explainer Augmentation Analysis.}
We continue to evaluate the explainer component by directly varying its parameter $K$, namely the number of its generated explanations used for augmenting reasoning and hinted classifier training.
As shown in Figure~\ref{fig:ablation_study_k}, the model performance first increases as $K$ increases until it reaches about 3 to 4 and then starts decreasing.
Notably, the performance dropping rates vary across different evaluation metrics.
The model accuracy drops faster compared to its recall.
This is probably because the quality of generated explanations will decrease if we force the explainer to generate lots of explanations.
Finally, we can see that for a wide range of $K$, the performance of \Our is better than \OurNoEX, which further demonstrates the usefulness of free-text explanations.


\begin{table*}[t]
        \scalebox{0.995}{
        \begin{tabular}{p{17.2cm}}
            \toprule
            \textbf{Headline}: WWE SmackDown results - \textcolor{orange}{2/9/19} \\
            \textbf{Article}: \underline{\emph{title}}: WWE Friday Night SmackDown Results \textcolor{cyan}{(3/26/21)} \underline{\emph{passage}}: WWE Friday Night SmackDown Results \textcolor{cyan}{March 26, 2021} Report by Lovell Porter for Wrestlezone.com You can also participate via social media by using the \#WZChat hashtag to voice your thoughts on tonight's show. We want you to share our exclusive coverage page ...\\
            \textbf{Human rated class}: ``Contradict'' \\
            \textbf{Human provided explanation}: ``the date in the headline different from the one appearing in the article.'' \\
            \textbf{\Our output}: ``Contradict because conflicting dates - \textcolor{cyan}{2021} vs \textcolor{orange}{2019}.'' \\
            
            \midrule
            \textbf{Headline}: SpaceX delays launch of \textcolor{orange}{Starlink satellites} \\
            \textbf{Article}: \underline{\emph{title}}: This Could Be a Big Reason Why SpaceX Is Delaying Its Starlink \textcolor{cyan}{IPO} \underline{\emph{passage}}: here is a lot of interest in investing in SpaceX, but an \textcolor{cyan}{IPO} making the company available to everyday investors is likely still a long way off. CEO Elon Musk has said SpaceX internet service provider subsidiary Starlink is much closer to ...\\
            \textbf{Human rated class}: ``Entail'' \\
            \textbf{Human provided explanation}: ``'' (\ie, rater does not give any explanation) \\
            \textbf{\Our output}: ``Contradict because \textcolor{cyan}{IPO} is missing in the headline which makes it misleading." \\
            
            \midrule
            \textbf{Headline}: Lockdown at 16 hotspots in Thane till \textcolor{orange}{March 31} \\
            \textbf{Article}: \underline{\emph{title}}: Maharashtra: Thane Municipal Corporation does U-turn on lockdown in 16 hotspots \underline{\emph{passage}}: HANE: The Thane Municipal Corporation (TMC) in Maharashtra, on Tuesday evening, did a complete U-turn on its Monday announcement of lockdown in 16 hotspots in the district. The TMC issued fresh guidelines tapering restrictions to the respective floors of a building or wings depending on the caseload, till \textcolor{cyan}{March end}, officials said.... \\
            \textbf{Human rated class}: ``Entail'' \\
            \textbf{Human provided explanation}: ``'' (\ie, rater does not give any explanation) \\
            \textbf{\Our output explanation}: ``Contradict because till \textcolor{orange}{March 31} is unsupported.'' \\
            
            \bottomrule
        \end{tabular}
        }
    \vspace{0.2cm}
    \caption{\textbf{\Our output case studies. We use two colors to highlight the key information in headlines and articles.}}
    \label{table:hallucination_cases}
\vspace*{-1.0em}
\end{table*}


\begin{table*}[t]
        \scalebox{0.995}{
        \begin{tabular}{p{17.2cm}}
            \toprule
            \textbf{Headline}: Kyle takes his own life in Hollyoaks \\
            \textbf{Article}: \underline{\emph{title}}: Does Kyle die in Hollyoaks? \underline{\emph{passage}}: Hollyoaks will tackle the subject of male suicide this week though a hard-hitting storyline featuring Kyle Kelly (Adam Rickitt). ... However, last week, conflict ensued between the friends after Kyle caused a devastating car crash while on drugs ...\\
            \textbf{Human rated class}: ``Contradict'' \\
            \textbf{Human provided explanation}: ``'' (\ie, rater does not give any explanation) \\
            \textbf{\Our explainer output 1}: ``Kyle is in a car crash so he doesn't take his own life.'' \\
            \textbf{\Our explainer output 2}: ``Kyle is in a car crash, not taking his own life.'' \\
            \textbf{\Our explainer output 3}: ``Kyle does not die in Hollyoaks. The show is about male suicide.'' \\
            \bottomrule
        \end{tabular}
        }
    \vspace{0.2cm}
    \caption{\textbf{Case study of the explainer module in our \Our framework.}}
    \label{table:hallucination_explainer_cases}
\vspace*{-1.0em}    
\end{table*}

\smallskip
\noindent \textbf{4. Case Studies.}
Table~\ref{table:hallucination_cases} shows some \Our output examples. More case studies are presented in Appendix~\ref{app:news_dataset_more_cases}.
First, we observe that \Our can generate high-quality human-readable explanations to justify its prediction.
In the first example, the model output explanation ``conflicting dates - 2021 vs 2019.'' captures the key difference between the headline and the article, and closely resembles the human written explanation ``the date in the headline different from the one appearing in the article''.

Second, we can see that \Our is able to help us identify potential labeling errors.
Take the second example as one case, the rater mistakenly labels it as an "Entail" case but in fact it should be misleading because the headline suggests the Starlink satellites launch is delayed but the article is about the delay of Starlink IPO. 
We can capture this error based on the model output explanation ``IPO is missing in the headline which makes it misleading''.

Moreover, we can see that the generated explanation enables us to understand why the model makes a certain mistake.
As shown by the last example in Table~\ref{table:hallucination_cases}, the headline is indeed supported by the news article but our \Our predicts it to be a contradiction case because the ``till March 31'' is not supported by the article.
Diving into the news article, we can see the ``till March 31'' information is referred to as ``till March end'' in the middle part of the article's main passage.
The model fails to recognize that the ``March end'' is synonymous with ``March 31'' and thus makes the wrong prediction.
This observation can motivate researchers to later study how to further increase the hallucination accuracy by improving the model's temporal reasoning ability.

Finally, Table~\ref{table:hallucination_explainer_cases} shows one example where the original curator does not provide any rating explanations other than the binary class label while our \Our explainer component can successfully generate valid explanations for the rated class.

\subsection{Zero-shot Hallucination Detection}


\begin{table*}[t]
        \scalebox{0.98}{
        \begin{tabular}{p{17.2cm}}
            \toprule
            \midrule
            \textbf{Dataset}: \colorbox{yellow}{QAGS} \\
            \textbf{Headline}: The number of illegal immigrants detained or arrested at the \textcolor{orange}{uk's busiest port} has more than doubled in a year, figures show. \\
            \textbf{Article}: There has been a large increase in the number of illegal immigrants detained or arrested by police at \textcolor{cyan}{dover}. Some 563 were held or arrested by \textcolor{cyan}{port of dover} police officers last year - up from 148 in 2013, a bbc radio kent freedom of information request shows. ... \\ 
            \textbf{Human rated class}: ``Contradict'' \\
            \textbf{\Our output}: ``Contradict because The number of illegal immigrants detained or arrested at \textcolor{cyan}{dover} - \textcolor{orange}{one of the uk's busiest ports} - has more than doubled, from 148 to 563.'' \\

            \midrule
            \textbf{Dataset}: \colorbox{yellow}{Vitamin-C} \\
            \textbf{Headline}: Averaged over \textcolor{teal}{more than 20 critics} , On the Basis of Sex scored \textcolor{teal}{less than 62 \%} . \\
            \textbf{Article}: On Metacritic , the film has a weighted average score of \textcolor{teal}{61 out of 100} , based on \textcolor{teal}{21 critics} , indicating ``generally favorable reviews'' . \\
            \textbf{Human rated class}: ``Entail'' \\
            \textbf{\Our output}: ``Entail because The film scored \textcolor{teal}{61 out of 100} , which is \textcolor{teal}{less than 62 \%} .'' \\
            \bottomrule
        \end{tabular}
        }
    \vspace{0.2cm}
    \caption{\textbf{\Our output case studies on TRUE benchmark datasets. If the article and the headline is contradictory, we use two different colors to highlight the key differences. Otherwise, we use one single color to underscore the shared key information.}}
    \label{table:true_hallucination_cases}
\vspace*{-0.5em}
\end{table*}

\subsubsection{Datasets.}
We further evaluate the zero-shot performance of \Our when no in-domain training data is provided. 
Specifically, we adopt the four summarization hallucination detection datasets: MNBM~\cite{MNBM}, FRANK~\cite{FRANK}, QAGS~\cite{QAGS}, SummEval~\cite{SummEval} and two fact verification datasets: FEVER~\cite{FEVER}, Vitamin-C~\cite{VitaminC} in the TRUE benchmark~\cite{Honovich2022TRUERF}.
Each dataset contains a set of $\langle$target text, grounding text, hallucination label$\rangle$ triplets where the binary label indicates if the target text is hallucinated based on the grounding text.
In the following experiments, we treat the target text as the ``headline'' and the grounding text as the ``article''.
More dataset details are available in Appendix~\ref{app:true_dataset_details}.

\begin{table}[!t]
\centering
    \scalebox{0.98}{
        \begin{tabular}{l|cc|cc}
            \toprule
            Datasets & Q2 & ANLI & \OurNoEX & \Our \\
            \midrule
            MNBM~\cite{MNBM}        & 66.5 & 66.7 & 73.8 & \bf 75.4 \\
            FRANK~\cite{FRANK}       & 82.9 & \bf 83.5 & 81.3 & 83.3 \\
            QAGS~\cite{QAGS}        & 78.3 & 75.3 & 76.6 & \bf 78.4 \\
            SummEval~\cite{SummEval}    & 77.3 & 72.9 & 85.6 & \bf 87.0 \\ 
            \midrule
            FEVER~\cite{FEVER}       & 82.7 & \bf 90.2 & 87.4 & 88.3 \\
            Vitamin-C~\cite{VitaminC}   & 75.7 & 74.7 & 84.8 & \bf 85.1 \\
            \midrule
            Average & 77.23 & 77.21 & 81.58 & \bf 82.91 \\
            \bottomrule
        \end{tabular}
    }
    \vspace*{0.2cm}
    \caption{Accuracy results on the TRUE datasets~\cite{Honovich2022TRUERF}.}
    \label{table:true_main_results}
\end{table}

\subsubsection{Compared Methods.}
We compare our \Our framework and its variant \OurNoEX with two best-performing methods in the original TRUE paper: 
(1) \textbf{ANLI} which, similar to our approach, first trains a T5-11B model using the ANLI dataset~\cite{Nie2020AdversarialNA} and then directly applies the learned model to detect the hallucinations, and 
(2) \textbf{Q2}~\cite{Q2} which first uses a question generation module to generate questions with answer spans from the target text and then applies a question answering (QA) model on the grounding text to answer the above-generated questions.
Finally, it computes the overlap between each true answer span and its corresponding QA model output answer span and outputs the final hallucination score.

\subsubsection{Experiment Settings.}
As no training example is provided in the TRUE benchmark, we reuse the \Our checkpoint after the NLI-based pretraining step and directly conduct the inference step of \Our on all tested datasets.
For fair comparisons, we follow the previous practices~\cite{Honovich2022TRUERF} to directly tune the binary cutoff threshold on the development set and report the best performance (in terms of accuracy) of all baseline methods in the original paper.

\subsubsection{Experiment Results.}
Table~\ref{table:true_main_results} shows the overall results on all six evaluated datasets.
We can see that both \Our and \OurNoEX can outperform the previous best methods and our \Our framework achieves new state-of-the-art results.
Moreover, by comparing \Our with \OurNoEX, we observe that adding explanation information is particularly useful in the zero-shot transfer learning setting.
We also demonstrate that \Our can augment the TRUE benchmark by providing interesting and insightful free-text explanations for the existing labels. 
As shown in Table~\ref{table:true_hallucination_cases}, \Our generates high quality human-readable explanations to explain its prediction results.
Take the case from the QAGS dataset as an example, \Our's output explanation captures the subtle semantic difference between \mquote{uk's busiest port} and \mquote{dover, one of the uk's busiest ports} and justifies why it makes the ``Contradict'' prediction.
Similarly, in the example from Vitamin-C dataset, \Our reiterates the fact that \mquote{scored 61 out of 100} (from the news article) implies \mquote{less than 62\%} (mentioned in the headline) and thus the headline is supported by the article.
More case studies are presented in Appendix~\ref{app:true_dataset_more_cases}.

\section{Conclusions and Future Work}\label{sec:conclusion}
This paper studies how to automatically detect news headline hallucinations with a limited amount of labeled data.
We propose a novel \Our framework which adapts knowledge from public NLI datasets into the news domain and generates natural language explanations to justify its prediction results.
Extensive experiments on one newly collected dataset and six public datasets demonstrate that \Our can accurately identify hallucinated news headlines along with high-quality human-readable explanations.

As a first-punch solution for detecting news headline hallucinations, we believe \Our can be improved in many ways. 
Interesting future directions include: 
(1) utilizing the validation set to learn a better combiner that better aggregates the predictions results from the reasoning classifier and the hinted classifier, 
(2) incorporating large language models (\eg, GPT-3, PaLM, ChatGPT) into \Our for better zero- and few-shot performance, 
(3) expanding the scope of \Our to the multilingual setting for detecting international news headline hallucinations,
(4) formatting the \Our output explanations to increase their readability, 
(5) enforcing the \Our output explanation itself to be entailed by the original news article and headline, and
(6) extending \Our to resolve multi-document headline hallucination problems where the headline is generated from multiple documents and we need to predict if it is hallucinated based on a whole set of documents.


\bibliographystyle{ACM-Reference-Format}
\bibliography{ref}

\clearpage
\appendix

\section{Experiment Details on News Hallucination Detection Dataset}\label{app:exp_details}

For all compared methods, we tune their hyper-parameters using the validation set, select the best ones, and report the corresponding results on the test set. 
Specifically, we have: for SVM\footnote{\small \url{https://scikit-learn.org/stable/modules/generated/sklearn.svm.SVC.html}}, we use the RBF kernel with C=0.1 and degree=4; for XGBoost\footnote{\small \url{https://xgboost.readthedocs.io/en/stable/index.html}}, we select gamma=1.0, max\_depth=3, min\_child\_weight=1, subsample=1.0, colsample\_bytree=0.5, and n\_estimators=30; for BERT$_{base}$\footnote{\small \url{https://github.com/tensorflow/models/tree/master/official/nlp}} and T5$_{xxl}$\footnote{\small \url{https://github.com/google-research/t5x}} methods, we select batch\_size=64 and learning\_rate=1e-3.
Both methods use a constant learning rate scheduler and are trained for 10k steps with 1k warmup steps. 
For our \Our framework and its variants, during the NLI-based pretraining stage, we choose batch\_size=128, constant learning\_rate=1e-3, and the number of explainer generated explanations $K=1$. During the domain fine-tuning stage, we select batch\_size=64, constant learning\_rate=1e-3, and the number of explainer generated explanations $K=3$. 
Both stages are trained for 10k steps with 1k warmup steps.
Finally, we train BERT$_{base}$, T5$_{xxl}$, and our models on TPU v3.

\section{TRUE Benchmark Datasets Statistics}\label{app:true_dataset_details}

Table~\ref{table:true_data_statistics} lists the statistics of TRUE benchmark datasets.

\begin{table}[!h]
\centering
    \scalebox{0.998}{
        \begin{tabular}{l|c}
            \toprule
            Datasets & \# of Examples \\
            \midrule
            MNBM~\cite{MNBM}        & 2,500 \\
            FRANK~\cite{FRANK}      & 671 \\
            QAGS~\cite{QAGS}        & 239 \\
            SummEval~\cite{SummEval} & 1,600 \\
            \midrule
            FEVER~\cite{FEVER}       & 18,209 \\
            Vitamin-C~\cite{VitaminC}   & 63,054 \\
            \bottomrule
        \end{tabular}
    }
    \vspace*{0.2cm}
    \caption{Statistics of TRUE Benchmark Datasets.}
    \label{table:true_data_statistics}
\end{table}

\section{Case Studies on News Hallucination Detection Dataset}\label{app:news_dataset_more_cases}

Table~\ref{table:hallucination_more_cases} lists case studies on our news hallucination detection dataset.


\begin{table*}[!h]
        \scalebox{0.96}{
        \begin{tabular}{p{17.2cm}}
            \toprule
            \textbf{Headline}: WWE SmackDown results - \textcolor{orange}{2/9/19} \\
            \textbf{Article}: \underline{\emph{title}}: WWE Friday Night SmackDown Results \textcolor{cyan}{(3/26/21)} \underline{\emph{passage}}: WWE Friday Night SmackDown Results \textcolor{cyan}{March 26, 2021} Report by Lovell Porter for Wrestlezone.com You can also participate via social media by using the \#WZChat hashtag to voice your thoughts on tonight's show. We want you to share our exclusive coverage page ...\\
            \textbf{Human rated class}: ``Contradict'' \\
            \textbf{Human provided explanation}: ``the date in the headline different from the one appearing in the article.'' \\
            \textbf{\Our output}: ``Contradict because conflicting dates - \textcolor{cyan}{2021} vs \textcolor{orange}{2019}.'' \\
            
            \midrule
            \textbf{Headline}: SpaceX delays launch of \textcolor{orange}{Starlink satellites} \\
            \textbf{Article}: \underline{\emph{title}}: This Could Be a Big Reason Why SpaceX Is Delaying Its Starlink \textcolor{cyan}{IPO} \underline{\emph{passage}}: here is a lot of interest in investing in SpaceX, but an \textcolor{cyan}{IPO} making the company available to everyday investors is likely still a long way off. CEO Elon Musk has said SpaceX internet service provider subsidiary Starlink is much closer to ...\\
            \textbf{Human rated class}: ``Entail'' \\
            \textbf{Human provided explanation}: ``'' (\ie, rater does not give any explanation) \\
            \textbf{\Our output}: ``Contradict because \textcolor{cyan}{IPO} is missing in the headline which makes it misleading." \\
            
            \midrule
            \textbf{Headline}: OnePlus 8T Cyberpunk 2077 Edition \textcolor{orange}{launched} \\
            \textbf{Article}: \underline{\emph{title}}: OnePlus 8T Cyberpunk 2077 Edition \textcolor{cyan}{unveiled} \underline{\emph{passage}}: OnePlus has \textcolor{cyan}{unveiled} a new special edition smartphone, the OnePlus 8T Cyberpunk 2077 Edition and the handset is now available to pre-order in China. The new OnePlus 8T Cyberpunk 2077 Edition ... \\
            \textbf{Human rated class}: ``Entail'' \\
            \textbf{Human provided explanation}: ``'' (\ie, rater does not give any explanation) \\
            \textbf{\Our output}: ``Contradict because \textcolor{orange}{launched} VS \textcolor{cyan}{unveiled}.'' \\
            
            \midrule
            \textbf{Headline}: Lockdown at 16 hotspots in Thane till \textcolor{orange}{March 31} \\
            \textbf{Article}: \underline{\emph{title}}: Maharashtra: Thane Municipal Corporation does U-turn on lockdown in 16 hotspots \underline{\emph{passage}}: HANE: The Thane Municipal Corporation (TMC) in Maharashtra, on Tuesday evening, did a complete U-turn on its Monday announcement of lockdown in 16 hotspots in the district. The TMC issued fresh guidelines tapering restrictions to the respective floors of a building or wings depending on the caseload, till \textcolor{cyan}{March end}, officials said.... \\
            \textbf{Human rated class}: ``Entail'' \\
            \textbf{Human provided explanation}: ``'' (\ie, rater does not give any explanation) \\
            \textbf{\Our output explanation}: ``Contradict because till \textcolor{orange}{March 31} is unsupported.'' \\
            
            \bottomrule
        \end{tabular}
        }
    \vspace{0.2cm}
    \caption{\textbf{\Our output case studies. We use two colors to highlight the key information in headlines and articles.}}
    \label{table:hallucination_more_cases}
\vspace*{-1.0em}
\end{table*}

\section{Case Studies on TRUE Benchmark Datasets}\label{app:true_dataset_more_cases}

Table~\ref{table:true_hallucination_more_cases} lists case studies on TRUE benchmark datasets.


\begin{table*}[!h]
        \scalebox{0.99}{
        \begin{tabular}{p{17.2cm}}
            \toprule
            \textbf{Dataset}: \colorbox{yellow}{MNBM} \\
            \textbf{Headline}: singer rod temperton, who wrote the hit album michael jackson, has died at the age of \textcolor{orange}{89}. \\
            \textbf{Article}: Temperton died in London last week at the age of \textcolor{cyan}{66} after ``a brief aggressive battle with cancer'', Jon Platt of Warner/Chappell music publishing said. Temperton's other hits included Off The Wall and Baby Be Mine for Jackson and Boogie Nights for his band Heatwave. Chic guitarist Nile Rodgers was among those paying tribute, tweeting: ... \\ 
            \textbf{Human rated class}: ``Contradict'' \\
            \textbf{\Our output}: ``Contradict because he did die at the age of \textcolor{cyan}{66}, not \textcolor{orange}{89}.'' \\

            \midrule
            \textbf{Dataset}: \colorbox{yellow}{FEVER} \\
            \textbf{Headline}: Heroes had \textcolor{teal}{a season with over 20 episodes}. \\
            \textbf{Article}: Heroes is an American psychological thriller science fiction television drama series created by Tim Kring that appeared on NBC for four seasons from September 25 , 2006 through February 8 , 2010 . The critically acclaimed \textcolor{teal}{first season had a run of 23 episodes} and garnered an average of 14.3 million viewers in the United States , ... \\
            \textbf{Human rated class}: ``Entail'' \\
            \textbf{\Our output}: ``Entail because The first season had \textcolor{teal}{a run of 23 episodes}.'' \\
            
            \bottomrule
        \end{tabular}
        }
    \vspace{0.2cm}
    \caption{\textbf{\Our output case studies on TRUE benchmark datasets. If the article and the headline is contradictory, we use two different colors to highlight the key differences. Otherwise, we use one single color to underscore the shared key information.}}
    \label{table:true_hallucination_more_cases}
\vspace*{-0.5em}
\end{table*}
\end{document}